\documentclass[letterpaper]{article}
\usepackage{uai2019}
\usepackage[margin=1in]{geometry}
\usepackage{times}
\usepackage[utf8]{inputenc}
\usepackage[english]{babel}
\usepackage{graphicx}
\usepackage{tikz}
\usetikzlibrary{automata,positioning}
\usepackage{verbatim}
\usepackage{enumitem}
\usepackage{amsfonts}
\usepackage{algorithm}
\usepackage{float}
\usepackage{subcaption}
\usepackage[noend]{algpseudocode}
\usepackage{booktabs}
\usepackage{xr}

\newcommand\ci{\perp\!\!\!\perp}
\newcommand{\gee}{\mathcal{G}}
\newcommand{\emm}{\mathcal{M}}

\newcommand{\bface}[1]{\textrm{\textbf{#1}}}
\newcommand{\enetwork}{\mathcal{E}_\mathcal{N}}
\newcommand{\eproto}{\mathcal{E}_{\textrm{proto }\mathcal{N}}}

\usepackage{amsthm}
\newtheorem{lemma}{Lemma}
\newtheorem{theorem}{Theorem}
\newtheorem{corollary}{Corollary}[theorem]
\theoremstyle{remark}

\theoremstyle{definition}

\usepackage{amsmath}
\usepackage{amssymb}
\usepackage[hidelinks]{hyperref}

\DeclareMathOperator{\pa}{pa}
\DeclareMathOperator{\nb}{nb}
\DeclareMathOperator{\bd}{bd}
\DeclareMathOperator{\cl}{cl}
\DeclareMathOperator{\doo}{do}
\DeclareMathOperator{\ant}{ant}
\DeclareMathOperator{\loc}{loc}
\DeclareMathOperator*{\argmax}{argmax}

\newenvironment{thma}[1]{\par\noindent{\bf Theorem #1\ }\em}{\em}
\newenvironment{lema}[1]{\par\noindent{\bf Lemma #1\ }\em}{\em}
\newenvironment{cora}[1]{\par\noindent{\textbf{Corollary #1} }\em}{\em}

\title{Causal Inference Under Interference And Network Uncertainty}

\author{ {\bf Rohit Bhattacharya}
\And
{\bf Daniel Malinsky}  \\
Department of Computer Science          \\
Johns Hopkins University \\
\{rbhattacharya@, malinsky@, ilyas@cs.\}jhu.edu
\And
{\bf Ilya Shpitser}
}

\begin{document}
	
\maketitle

\begin{abstract}
  Classical causal and statistical inference methods typically assume the observed
  data consists of independent realizations.
  However, in many applications this assumption is inappropriate due to a
  network of dependences between units in the data.
  Methods for estimating causal effects have been developed in the setting where
  the structure of dependence
  between units is known exactly \cite{hudgens08, tchetgen2012causal, ogburn14},
  but in practice there is often substantial uncertainty about the precise network structure.
  This is true, for example, in trial data drawn from vulnerable communities where social
  ties are difficult to query directly.
  In this paper we combine techniques from the structure learning and interference literatures in causal inference, proposing a general method for estimating causal effects
  under data dependence when the structure of this dependence is not known a priori.
  We demonstrate the utility of our method on
  synthetic datasets which exhibit network dependence.
\end{abstract}

\section{INTRODUCTION}
In many scientific and policy settings, research subjects do not exist in isolation but in
interacting networks.
For instance, data drawn from an online social network will exhibit
\emph{homophily} (friends are similar, because they are friends), and
\emph{contagion} (friends may causally influence each other)
\cite{lewis2012social,shalizi2011homophily,kramer2014experimental,sherman2018identification}.  Similarly, vaccinating some subset of
a population may confer immunity to the entire population -- a
well-documented phenomenon known as \emph{herd immunity} in infectious
disease epidemiology.  This implies that a treatment given to one unit
affects outcomes for another.  Finally, resource constraints in
allocation problems may also induce data dependence.


In the context of causal inference, methods for dealing with data
dependence are developed under the heading of
\emph{interference} \cite{sobel2006randomized,hong2006evaluating,rosenbaum2007interference,hudgens08,tchetgen2012causal,shalizi2011homophily,ogburn14, auto-g}.
Most such work assumes the structure of the dependence (which units
depend on which others, and how) is known precisely.  For example,
\cite{tchetgen2012causal} assumes units in the data may be organized
into equal sized blocks, where units within a block are pairwise
dependent and units across blocks are not.  Some work makes
alternative assumptions, e.g., \cite{auto-g} assumes that blocks are
drawn from a known random field.

In many applications, the network inducing dependence between units may not be
known exactly.  For instance, in vulnerable, stigmatized, or isolated communities
(such as groups of drug users, or remote villages),
we may have no way of reconstructing the precise social ties between individuals.
Some online databases of social media users may be anonymized, with friendship
ties deliberately omitted. There has been some work in such settings that
involves adapting the data collection method itself in order to
discover the underlying networks: e.g., snowball sampling in
\cite{crawford2017sampling} and \cite{bramoulle2016oxford}.  
Unfortunately, such study designs are not always possible
to arrange in advance, and most data available on networks of interacting units
is not collected under such designs.

While there is a rich literature on model selection from observational
data in the context of causal inference (e.g.,
\cite{spirtes00,chickering2002ges,shimizu14,peters14}), to our
knowledge all previous work has assumed the absence of interference.
We explore learning the dependence structure using
graphical model selection methods.  Techniques for structure
learning from probabilistic relational models are also related to this work
\cite{maier2013sound, lee2016learning}.

The contributions of our paper may be viewed in one of two ways.
From the point of view of causal inference under interference, our
paper contributes to methods for estimating causal effects when there is
substantial uncertainty about network structure.  
From the point of view of structure learning,
we introduce novel algorithms for model selection when
units are dependent due to a network, the structure of which is unknown.

\section{MOTIVATING EXAMPLE AND BACKGROUND ASSUMPTIONS}
To motivate our work, we discuss an example application.
Consider a public health program aimed at lowering the incidence of
blood-borne diseases such as HIV in at-risk individuals who are
addicted to heroin and share needles when injecting intravenously.
An example of such a program is described in
\cite{stancliff2003syringe}.  The program creates
pop-up clinics around the city where disposable needles are
distributed for free to individuals in need, but due to limited
resources only a limited number of individuals will actually
receive these needles. We would like to know, in this restricted resource
setting, if the use of disposable needles spreads amongst the rest of
the population.  Additionally, we would like to detect the phenomenon
of herd immunity, i.e.,\ whether some members of the population being
protected due to taking advantage of the clean needles confer this
protection to others who do not.

Data on heroin users was collected via such program, with users
arranged by neighborhood or municipality.  Users in different
neighborhoods are assumed independent, but users within the same
neighborhood are likely dependent.  This setting is known as \emph{partial
interference} \cite{hudgens08}.
For each individual $i$, data is collected on their use of disposable needles $A_i$,
their subsequent health outcome $Y_i$ (risk of obtaining blood-born disease),
along with a vector of pre-treatment covariates $\bface{L}_i = (L_{1,i},...,L_{p,i})$.
We may be interested in quantifying the causal effects of $A_i$ on $Y_j$,
for arbitrary $i$ and $j$ within a neighborhood, or network-averaged versions of such
effects \cite{ogburn14}.

We may assume that background knowledge or study design implies a ``known''
individual-level causal structure for each $i$, namely that 
$A_i \to Y_i$ and $A_i \leftarrow \bface{L}_i \to Y_i$,
but that we are uncertain about network ties among users.
One approach is to assume the least restrictive model, where all users in a
neighborhood are arbitrarily dependent. This would correspond to a complete network, where every pair of vertices is directly connected.
However, assuming a complete network when the true network is sparse ignores useful structure in the problem and leads to inefficient estimates of target quantities.  In addition, complete networks often lead to likelihoods that are intractable to evaluate.
An alternative is to a select a sparse network supported by the data.  In addition to enabling tractable and statistically efficient inference, such an approach may also rule out the presence of certain causal effects
without explicitly estimating them, if corresponding pathways are absent in the selected network.

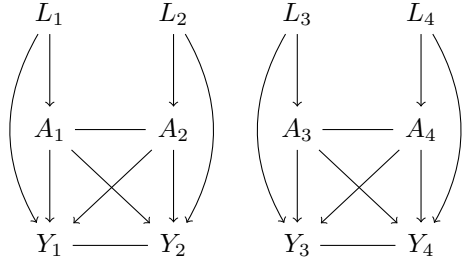
\begin{figure}[t]
	\centering
	\begin{tikzpicture}
	\node[]  (l1) {$L_1$};
	\node[right=of l1] (l2) {$L_2$};
	\node[right=of l2] (l3) {$L_3$};
	\node[right=of l3] (l4) {$L_4$};
	\node[below=of l1] (a1) {$A_1$};
	\node[below=of l2] (a2) {$A_2$};
	\node[below=of l3] (a3) {$A_3$};
	\node[below=of l4] (a4) {$A_4$};
	\node[below=of a1] (y1) {$Y_1$};
	\node[below=of a2] (y2) {$Y_2$};
	\node[below=of a3] (y3) {$Y_3$};
	\node[below=of a4] (y4) {$Y_4$};
	
	\draw[every loop]
	(y1) edge[-] node {} (y2)
	(y3) edge[-] node {} (y4)
	(a1) edge[-] node {} (a2)
	(a3) edge[-] node {} (a4)
	(l1) edge[] node {} (a1)
	(a1) edge[] node {} (y1)
	(l1) edge[bend right] node {} (y1)
	(l2) edge[] node {} (a2)
	(a2) edge[] node {} (y2)
	(l2) edge[bend left] node {} (y2)
	(l3) edge[] node {} (a3)
	(a3) edge[] node {} (y3)
	(l3) edge[bend right] node {} (y3)
	(l4) edge[] node {} (a4)
	(a4) edge[] node {} (y4)
	(l4) edge[bend left] node {} (y4)
	(a1) edge[] node {} (y2)
	(a2) edge[] node {} (y1)
	(a3) edge[] node {} (y4)
	(a4) edge[] node {} (y3)
	;
	\end{tikzpicture}
	\caption{A chain graph over three variables ($L, A,$ and $Y$) on 4 individuals, representing possible relationships between disposable needle use and risk of blood-borne disease among heroin-users.}
	\label{needles_network}
\end{figure}

As an example, if neighborhoods have $4$ units, we may aim to learn a
graphical model such as shown in Figure \ref{needles_network}.  This
model, containing both directed edges (representing direct causal
influences) and undirected edges (representing symmetric network
ties), is known as a chain graph model \cite{lauritzen1996}.  We
describe chain graphs in more detail below.  This model tells us that
we should expect some spread of disposable needle use from one unit to
another.  However, it also tells us that users in neighborhoods are
split into two non-interacting groups: $\{ 1, 2 \}$ and $\{ 3, 4 \}$.
This implies the absence of contagion from one group to another. In
addition, the conditional independences among units implied by this split suggests that
contagion effects within groups may be estimated more efficiently as
compared to a statistically saturated model, with a complete network
across units.

The algorithms we propose are consistent (in the sense that they asymptotically converge on the true model) under a set of assumptions which we now informally summarize. We assume the true data-generating process corresponds perfectly (satisfying Markov and faithfulness conditions) to some unknown chain graph, with two restrictions: (1) the unit-level graph is known, reflecting the aforementioned causal ordering between pre-treatment covariates, treatment variables, and outcomes; and (2) the graph respects what we later call \emph{tier symmetry}, which restricts connections between variables at the same ``tier'' in the causal ordering to be symmetric. We assume the data is distributed with some (known) likelihood in the exponential family, as well as some weak statistical regularity conditions. We also present algorithms that make an additional simplifying assumption on the graphical structure -- namely that influence between units is the same for all unit pairs -- but such an assumption is not strictly necessary for consistency.

We begin by describing some technical preliminaries, including chain graph models,
causal inference, and graphical model selection. Then we present algorithms to learn graphical models
of the sort shown in Figure \ref{needles_network}, before estimating causal effects.

\section{PRELIMINARIES}
\subsection{Graphical Terminology}
Chain graphs (CGs) are a class of mixed graphs containing directed
($\rightarrow$) and undirected ($-$) edges, such that it is impossible
to create a directed cycle by orienting any combination of the undirected edges
\cite{lauritzen1996}. A CG with no undirected edges is a directed
acyclic graph (DAG). A CG with no directed edges is an undirected graph (UG) 
Vertices of a graph are denoted by capital letters
(e.g.\ $A$), and they correspond to random variables.  We use boldface
(e.g.\ $\bface{A}$) to denote sets of vertices or sets of random
variables. Lowercase letters denote specific values of random
variables (e.g.\ $a$) or sets of values (e.g.\ $\bface{a}$). We use
$\bface{V}$ and ${\cal E}$ to denote the set of all vertices and edges
in a graph $\gee$, respectively.

For a subset of vertices $\bface{A} \subseteq \bface{V}$ we define the
induced subgraph $\gee_\bface{A}$ to be the graph with vertices
$\bface{A}$ and edges of $\gee$ that have both endpoints in
$\bface{A}$. A block $\bface{B}$ is defined as a maximal set of
vertices such that every vertex pair in $\gee_\bface{B}$ is connected
by an undirected path.  The set of blocks in a CG ${\cal G}$, denoted by ${\cal B}({\cal G})$, partitions
the vertices in ${\cal G}$.
A clique $\bface{C}$ is defined as a maximal set of vertices
that are pairwise connected by undirected edges.  A clique in a CG is
always a subset of some block $\bface{B}$. We denote the set of all cliques
in an UG ${\cal G}$ by ${\cal C}({\cal G})$.

For a graph $\gee$ and vertex $V \in \bface{V}$ we define some
standard vertex sets as follows: the set of parents $\pa_\gee (V)
\equiv \{ V' : V' \rightarrow V \textrm{ in } \gee \}$; the set of
neighbors $\nb_\gee (V) \equiv \{ V' : V' - V \textrm{ in } \gee \}$;
the boundary $\bd_\gee (V) \equiv \pa_\gee (V) \cup \nb_\gee (V)$; and the
closure $\cl_\gee (V) \equiv \bd_\gee (V) \cup V$.
These definitions generalize disjunctively to sets,
e.g. $\pa_{\cal G}(\bface{A}) \equiv \bigcup_{A \in \bface{A}} \pa_{\cal
  G}(A)$.  Note that for a block $\bface{B}$, $\bd_{\cal G}(\bface{B}) =
\bface{B} \cup \pa_{\cal G}(\bface{B})$.  Given a CG ${\cal G}$, define
the augmented graph ${\cal G}^a$ to be an UG constructed from ${\cal G}$ by replacing all directed edges with undirected edges and connecting all vertices in $\pa_{\cal G}(\bface{B})$ for every block $\bface{B}$ in ${\cal G}$ by undirected edges.

We will utilize chain graphs to represent both causal relationships and network dependence
among units 
that form a (``social'') network ${\cal N}$. The undirected network ${\cal N}$ is a graph (distinct from our CG of interest) where the vertices correspond to units (e.g.\ individuals $i,j,...$), not random variables. Units may be adjacent or non-adjacent in ${\cal N}$ based on whether they are ``friends'' or otherwise directly dependent.


For each unit $i$, we denote the unit-level variables for $i$ in the CG $\gee$ (e.g.,\ $L_i, A_i,$ and $Y_i$ in Figure \ref{needles_network}) by $\bface{V}_i$, and edges among those variables by
${\cal E}_i$.  Similarly, for a pair of units $i,j$ which are adjacent in ${\cal N}$, we represent the set of
edges from $\bface{V}_i$ to $\bface{V}_j$ (and vice versa) by ${\cal E}_{ij}$.
It is the presence of these edges that induces data dependence between $i$ and $j$ in our analysis.
The set of ${\cal E}_{ij}$ for all pairs $i,j$ adjacent in ${\cal N}$ (i.e., the set of all cross-unit edges) will be denoted by ${\cal E}_{\cal N}$.

\subsection{Chain Graph Models}
\label{subsec:cg-stat}

A statistical chain graph model associated with a LWF (Lauritzen-Wermuth-Frydenberg) chain graph ${\cal G}$ is a set of distributions that factorize as:
\begin{align} \label{eqn:cg-f}
&p(\bface{V}) = \prod_{\bface{B} \in {\cal B}({\cal G})} p(\bface{B} \mid \pa_{\cal G}(\bface{B})) \notag\\
&\text{and} \\
&p(\bface{B} \mid \pa_{\cal G}(\bface{B})) = 
	\frac{
	\prod_{ \{ \bface{C} \in {\cal C}(({\cal G}_{\bd_{\cal G}(\bface{B})})^a): \bface{C} \not\subseteq \pa_{\cal G}(\bface{B}) \} }
		\phi_\bface{C}(\bface{C})
	}{Z(\pa_{\cal G}(\bface{B}))}
		\notag
\end{align}
for each block $\bface{B}$ in ${\cal G}$, where $\phi_\bface{C}(\bface{C})$ is a clique potential function for a clique $\bface{C}$ in the UG
$({\cal G}_{\bd_{\cal G}(\bface{B})})^a$ defined as above and $Z(\pa_{\cal G}({\bf B}))$ is a normalizing function \cite{lauritzen1996}.

A CG without undirected edges is a DAG,
which has a simpler factorization: $p(\bface{V}) = \prod_{V \in \bface{V}} p(V \mid \pa_{\cal G}(V))$. If it is the case that for every
block $\bface{B}$ in CG ${\cal G}$, ${\cal G}_{\bd_{\cal G}(\bface{B})}$
has missing edges only among elements of $\pa_{\cal G}(\bface{B})$, then
$({\cal G}_{\bd_{\cal G}(\bface{B})})^a$ has a single clique
containing all elements in $\bd_{\cal G}(\bface{B})$. In other words,
the model corresponding to such a CG may be viewed as a DAG model with entire
blocks $\bface{B}$ acting as vertices in a DAG.


\subsection{Causal Models}

A causal model is a set of distributions over counterfactual
random variables, a.k.a.\ potential outcomes.  
For $Y \in
\bface{V}$ and $\bface{A} \subseteq \bface{V} \setminus Y$, the counterfactual
$Y(\bface{a})$ denotes the value of $Y$ when the ``treatment'' variables
$\bface{A}$ are fixed to values $\bface{a}$ by an intervention. Sometimes interventions are formalized by the `do'-operator: $\doo(\bface{a})$ denotes the assigment $\bface{a}$ to $\bface{A}$ \cite{pearl2009causality}. The counterfactual distribution corresponding to the intervention where $\bface{A}$ is set to $\bface{a}$ is written $p(Y(\bface{a}))$ or $p(Y|\doo(\bface{a}))$. 

A causal model of a DAG $\gee$ is a set of distributions defined on
counterfactual random variables $V(\bface{a})$ for each $V \in
\bface{V}$ and where $\bface{a}$ is a set of values for $\pa_\gee(V)$.
Equivalently, a causal model can be understood as the set of distributions induced by
a system of structural equations (one equation for each vertex) equipped with the $\doo(\cdot)$ operator \cite{pearl2009causality, richardson2013single}.
In a causal model of
a DAG $\gee$, all counterfactual distributions are \textit{identified},
i.e.,\ they can be expressed as functions of the observed data, by the
g-formula \cite{robins1986new}: 
\begin{equation*}
p(\bface{V} \setminus \bface{A} \mid \doo(\bface a)) = \prod_{V \in \bface V \setminus \bface A} p(V \mid \pa_\gee(V)) \bigg \rvert_{\bface A = \bface a}.
\end{equation*}
Counterfactual responses to interventions are often contrasted on a
mean difference scale under two possible interventions $\bface{a}$ and
$\bface{a}'$, representing cases and controls. For example, the
average causal effect (ACE) is given by $\mathbb{E}[Y({\bface{a}})] - \mathbb{E}[Y(\bface{a}')]$.

Causal models have been generalized from DAGs to CGs (details in the Supplement) and
yield the following generalization of the g-formula \cite{lauritzen2002chain}:
\begin{equation}
p(\bface{V} \setminus \bface{A} | \doo(\bface a)) = \prod_{\bface B \in \mathcal{B}(\gee)}
p(\bface B \setminus \bface A \mid \pa(\bface B), \bface B \cap \bface A) \bigg \vert_{\bface A = \bface a}.
\label{eqn:cg-g}
\end{equation}


\subsection{The Conditionally Ignorable Network Model and Network Causal Effects}
\label{subsec:net-cim}

For the purposes of this paper, we consider CGs decomposed into three disjoint sets of variables:
$\bface{L}$, representing vectors of baseline (pre-treatment) factors; $\bface{A}$, representing treatments;
and $\bface{Y}$, representing outcomes.  For each unit $i$, we assume
$\bface{L}_i \subseteq \pa_{\cal G}(A_i)$, and $\bface{L}_i \cup \{ A_i \} \subseteq \pa_{\cal G}(Y_i)$.
This represents a common assumption (which we call \emph{causal ordering})
in causal inference that for each unit
both baseline factors and treatment potentially affect the outcome, and that the
baseline factors also affect treatment assignment.  Here each unit has one treatment variable $A_i$, one outcome variable $Y_i$, and possibly many baseline variables $\bface{L}_i$. In
interference settings, it is standard to allow that variables for another unit $j$ may influence variables for
unit $i$.  In our case, there is a further complication: the precise nature of this influence is unknown.

This model implies, for positive $p(\bface{V})$, the following standard assumptions from the interference literature:
$\bface{Y}(\bface{a}) \ci \bface{A} \mid \bface{L}$ (network ignorability);
$p(\bface{a} \mid \bface{L}) > 0$ $\forall \bface{a}$ (positivity); and
$\bface{Y}(\bface{a}) = \bface{Y}$ if $\bface{A} = \bface{a}$ (consistency).
Under these assumptions, the joint counterfactual
outcome is identified, regardless of the underlyling network structure,
 as the following special case of (\ref{eqn:cg-g}):
$p(\bface{Y}(\bface{a})) = \sum_{\bface{L}} p(\bface{Y} \mid \bface{A}=\bface{a},\bface{L}) p(\bface{L})$.

Given a particular treatment assignment probability $\pi(\bface{A})$,
a number of causal effects of interest may be defined, see \cite{tchetgen2012causal}
for an extensive discussion.  In this paper, we
focus on a single effect, the \emph{population average overall
effect (PAOE)}, though our results generalize to any identified
causal effect of interest in network settings (for example, spillover effects).
Consider a block is of size $m$ and two fixed $\pi_1,\pi_2$ assignment probabilities. Then
the PAOE is defined as:
\begin{align}
\frac{1}{m} \sum_{i=1}^m \sum_{\bface{A}} \mathbb{E}[Y_i(\bface{A})] \{ \pi_1(\bface{A}) - \pi_2(\bface{A}) \}.
\label{eqn:paoe}
\end{align}
Under the aforementioned assumptions, this effect is identified by the following functional \cite{tchetgen2012causal}:
\begin{align}
\frac{1}{m} \sum_{i=1}^m \sum_{\bface{L},\bface{A}} \mathbb{E}[Y_i \mid \bface{A}, \bface{L}] p(\bface{L}) \{ \pi_1(\bface{A}) - \pi_2(\bface{A}) \}.
\label{eqn:paoe-id}
\end{align}

A number of estimation strategies for (\ref{eqn:paoe-id}) are possible
under various assumptions on network structure.  For example,
\cite{tchetgen2012causal} considered an inverse probability weighted
estimator.  In this paper, we use the auto-g-computation algorithm in
\cite{auto-g} to estimate the PAOE, which allows for arbitrary network structure; we describe this estimator in detail in the
Supplement.

\section{MODEL SELECTION FOR UNKNOWN NETWORKS}

We are interested in estimating causal effects like the PAOE under the
aforementioned assumptions, where there is uncertainty about the network structure.  We give a taxonomy of problems of this type, having different levels of
difficulty depending on the degree of uncertainty present.

%

The most general version of the problem occurs when neither the causal
structure of each unit, nor the network structure inducing dependence
between units, is known. In this case the problem reduces to a
structure learning problem for arbitrary chain graphs, as considered
in \cite{ma08} and \cite{pena14}.  We do not pursue this version of
the problem here for two reasons.  First, the causal structure for
each unit is often known due to background knowledge on temporal
ordering and study design, as is the case for our needle-dispensary
motivating example.  Second, model selection of arbitrary CGs is known
to be a very challenging problem which (in the worst case) may require large
sample sizes \cite{evans18}.

In many settings, the causal structure for each individual unit is known and is
typically assumed to be the same for every unit, i.e.,\ $\mathcal{E}_i =
\mathcal{E}_j$ for all $i$, $j$. The problem of model selection then
amounts to learning the structure of the connections between units
i.e., $\mathcal{E}_{ij}$ for all $i$, $j$. The search space for such a
problem, while much smaller than the general problem, is still
exponential.  For a block that contains $m$ units, there are
${m\choose 2}$ possible pairings of units, leading to $2^{m\choose 2}$
possible networks.  The number of possible valid chain graphs is even
larger, since units $i,j$ adjacent in a network could be connected in
a variety of ways via (undirected or directed) edges in ${\cal E}_{ij}$. Learning these
connections requires a search through all possible combinations of
edges that form ${\cal E}_{ij}$ such that the overall graph is a
CG.

We may restrict the problem further by requiring that the connections
between any two units, if present, are \textit{homogenous},
meaning that dependence between any two units,
if it exists, arises in the same way. Formally, we define homogeneity such that,
for all pairs $(i, j), (k, l) \in \mathcal{N}$, $\mathcal{E}_{ij} =
\mathcal{E}_{kl}$. 
Notice that the space of homogenous networks is still fairly
large. The problem may be made more tractable by one of the following
two assumptions.  We may assume the existence of network
connections is known, but that their types are unknown, i.e., we know $\cal{N}$
and would like to learn $\mathcal{E}_{ij}$.  Alternatively, we may assume
we know how two adjacent units are connected, but not which pairs are
adjacent, i.e.,\ we know $\mathcal{E}_{ij}$ and would like to learn $\cal{N}$. We may also have no such background knowledge.
In the following, we present algorithms for both homogenous and heterogenous settings.

Throughout, we make an assumption which we call
\emph{tier symmetry}, which is commonly made implicitly or explicitly
in the interference literature \cite{tchetgen2012causal, auto-g}. That
is, we require connections between variables in the same ``tier'' of
causal ordering to represent symmetric relations between
the variables. This restricts edges $L_i - L_j$, $A_i - A_j$, and $Y_i
- Y_j$ to always be undirected. Also it is natural to extend
the known causal ordering of variables to connections between units: while we allow for e.g., $A_i \rightarrow Y_j$, the reverse, $Y_j
\rightarrow A_i$ is ruled out.  Finally, we rule out the existence of
undirected edges connecting variables across tiers, e.g, edges of the
form $A_i - Y_j$, since the existence of such edges, coupled with our
causal ordering assumption, leads to graphs which are not CGs.

Before presenting algorithms to address the above taxonomy of
problems, we introduce some necessary concepts from the graphical
model selection literature.

\subsection{Markov Properties and Faithfulness}

If $p(\bface{V})$ is a positive distribution,
the factorization (\ref{eqn:cg-f}) is equivalent to a global Markov property
which relates certain graphical separation facts in the CG ${\cal G}$ (given by the c-separation criterion) to conditional independence relations in $p(\bface{V})$; see \cite{lauritzen1996} for precise definitions. 
In what follows, we make the \emph{faithfulness} assumption, which is
the converse of the global Markov property: if $(\bface{A} \ci \bface{B}
\mid \bface{C})$ in $p(\bface{V})$, then $\bface{A}$ is c-separated from
$\bface{B}$ given $\bface{C}$ in ${\cal G}$. This is directly analogous to the faithfulness assumption made when selecting DAG models from data by constraint-based or score-based methods \cite{spirtes00, chickering2002ges}.

\subsection{Model Scores and the Pseudolikelihood}

In this paper, we will learn the structure of the network using a score-based approach
to model selection.  Score-based methods proceed by choosing the graph
(from among some space of candidates) that optimizes a model score.
Exhaustive model search is typically infeasible, so
it is popular to employ greedy methods that optimize only ``locally,'' that is, they traverse the space of candidate graphs considering only single-edge additions and deletions. 
Under some conditions, such greedy procedures can be shown to asymptotically converge to
the globally optimal model \cite{chickering2002ges}. Scores used for
greedy search typically satisfy three properties that are
sufficient for finding the globally optimal model: decomposability,
score-equivalence, and consistency.

A score is said to be decomposable if it can be written as a sum of
local contributions, each a function of one vertex and its
boundary. A score is said to be score-equivalent if two Markov
equivalent graphs (i.e., graphs that imply the same set of conditional
independences by the global Markov property) yield the same score.  A
score is said to be consistent if, as the sample size goes to
infinity, the following two conditions hold.  First, when two models
both contain the true generating model, the model of lower dimension
will have a better score.  Second, when one model contains the true
model and another does not, the former will have a better score.

A popular score satisfying these properties for model selection among DAG models is the Bayesian Information Criterion (BIC)
\cite{schwarz1978}.  Given a $d$-dimensional data set $\bface{D}$ of size $n$ and model likelihood
${\cal L}(\bface{D}; {\cal G}) \equiv \prod_{i=1}^n p(x_{1,i}, \ldots, x_{d,i}; {\cal G})$,
the BIC is given by $2 \ln {\cal L}(\bface{D}; {\cal G}) - k \ln (n)$ where $k$ is model
dimension.

For CG models, the BIC is only decomposable for blocks, not for variables within the block.
In addition, the score is not easy to evaluate.  Both of these issues arise due to the presence
of normalizing functions in the likelihood.  Here, we present an alternative score which
avoids some of these problems, based on the \emph{pseudolikelihood function} \cite{besag1974}:
\[
{\cal PL}(\bface{D};{\cal G}) \equiv \prod_{i=1}^n \prod_{j=1}^d p(x_{j,i} \mid \bface{x}_{-j,i}; {\cal G}),
\]
where $\bface{x}_{-j}$ is the vector $(x_1, \ldots, x_{j-1},x_{j+1},
\ldots, x_d)$.  We define a score based on the pseudolikelihood called
Pseudo-BIC (PBIC):
$2 \ln {\cal PL}(\bface{D}; {\cal G}) - k \ln (n)$.

We propose a greedy score-based model selection procedure based on the
PBIC score, which is consistent and obeys a weaker notion of decomposability
for exponential families, as we show below.
All proofs are deferred to the Supplement.

\begin{lemma}
With dimension fixed and sample size increasing to infinity,
the PBIC is a consistent score for curved exponential families whose natural parameter space $\Theta$ forms a compact set.
\label{pbic_consistency}
\end{lemma}
%


Decomposability of a scoring criterion makes greedy search a
practical procedure, by limiting the number of terms in the overall
score that need to be recomputed for each considered edge modification.
While the BIC score for DAG models
is decomposable,
the PBIC score for CG models is not.
Nevertheless, a weaker notion of decomposability holds, which implies that two CG
models that differ by a single edge differ by a subset of components of the score, which
we now describe. 
Consider a candidate edge between $V_i$ and $V_j$ in a CG ${\cal G}$.
Let $\bface B_{\loc}$ denote the block to which $V_j$ belongs when the edge is
directed $V_i \rightarrow V_j$, or to which $V_i$ and $V_j$
belong when the edge is undirected $V_i - V_j$. We use $\loc(V_i, V_j; \gee)$ to
denote a set of vertices called the \emph{local set}, defined as:
\begin{align*}
\bigcup\limits_{\bface{C}}\{&
\bface C \in \mathcal{C}((\gee_{\bd_\gee(\bface B_{\loc})})^a) : 
V_i,V_j \in \bface{C} \not\subseteq \pa_{\cal G}(\bface{B}_{\loc})\}.
\end{align*}
As we show, the score difference for graphs $\gee$ and $\gee'$ which
differ by a single edge can be written as the difference between
terms that involve only variables in the local set of $\gee$. The next result, 
and much subsequent discussion in the paper, is stated for conditional Markov random fields (MRFs). This is because statistical CG models can be equivalently described as sets of conditional MRF models. We elaborate on this relationship in the Supplement.  


\begin{lemma}
\label{locality}
Let $\gee$ and $\gee'$ be graphs which differ by a single edge between $V_i$ and $V_j$.
For conditional MRFs in the exponential family, the local score
difference between $\gee$ and $\gee'$ is given by:
$\sum_{V \in \loc(V_i, V_j; \gee) \cap \bface{B}_{\loc}} \{s_V\big(\bface{D}; \gee \big) - s_V\big(\bface{D}; \gee' \big)\},$ where
$s_V(.)$ denotes the component of the score for $V$.
\end{lemma}

Note that the above
definition of the local set may simplify further in certain special
cases of MRF models in the exponential family.  In particular, if we consider
an MRF that is multivariate normal, or a log linear discrete model with
only main effects and pairwise interactions, then the sum in Lemma 2 reduces to either a sum over elements $V_i$ and $V_j$ (for an undirected edge $V_i - V_j$) or only $V_j$ (for a directed edge $V_i \to V_j$).
We omit the straightforward proofs in the interest of space.  We will
not consider these special instances of the exponential family in
the remainder of this paper, but in the supplement we discuss 
the incurred computational costs 
for exponential families in general.

\subsection{Greedy Network Search}

While there exist numerous methods that take a pseudolikelihood-type
approach to model selection in UGs
\cite{ravikumar2010high,jalali2011learning,foygel2010extended,barber2015high},
these have been typically restricted to Ising or Gaussian 
models. Such methods involve a per-vertex neighbourhood selection
procedure using L1-regularized regression or the standard BIC, which
may yield self-inconsistent results (e.g., find that $V_i$ in $\nb (V_j)$ but not
vice versa).  Any resulting inconsistencies would need to be resolved post hoc through
union or intersection consolidation procedures. Methods that try to
enforce self-consistency
by explicitly maximizing the
pseudolikelihood with a regularization penalty are presented in
\cite{hofling2009estimation} and \cite{khare2015convex}, but are again
restricted to Ising and Gaussian graphical models. The properties of
the PBIC described in the previous section allow us to design
algorithms for greedy network search that are parallelizable, while
also generalizing to all exponential families and circumventing the
need for post hoc procedures. While our method covers a more general
class of models, it can be computationally expensive to calculate the
local scores at each step. A more efficient procedure is possible in
some subclasses (including Ising and Gaussian), where we can modify
our procedure into a ``forward-backward'' algorithm reminiscent of the GES
algorithm \cite{chickering2002ges}. 
Since our focus is on a general procedure for all exponential families,
we defer further discussion of these special cases to the Supplement.

\begin{algorithm}[t]
	\caption{\textproc{Greedy Network Search}$(\gee^{\textrm{init}}, \bface{D})$} \label{alg:greedy}
	\begin{algorithmic}[1]
		\State $\gee^* \gets \gee^{\textrm{init}}$
		\State $\textrm{score change} \gets \textrm{True}$
		\While{$\textrm{score change}$}
		\State $\textrm{score change} \gets \textrm{False}$
		\State $\mathcal{E}^*_{\mathcal{N}} \gets$ network ties in $\gee^*$
		\State $E_{max} \gets \argmax_{E \in \enetwork^*} \textrm{PBIC}(\bface{D}; \gee^* \setminus E)$
		\If{$\textrm{PBIC}(\bface{D}; \gee^* \setminus E_{max}) > \textrm{PBIC}(\bface{D}; \gee^*)$}
		\State $\gee^* \gets \gee^* \setminus E_{max}$
		\Comment{deleting edge $E_{max}$}
		\State $\textrm{score change} \gets \textrm{True}$
		\EndIf
		\EndWhile
		\State \textbf{return} $\cal{E}^*_{\cal N}$
	\end{algorithmic}
\end{algorithm}

We begin by describing a greedy search procedure that learns network
ties $\enetwork$, without imposing homogeneity. 
Model selection proceeds by solving 3 independent
sub problems: learning a Markov random field (MRF) over the
baseline covariates $\bface{L}$, learning a conditional MRF on the
treatments $\bface{A}$, and learning a conditional MRF on the outcomes
$\bface{Y}$. The resulting network ties learned from each of these,
are combined to produce the final result (Alg. \ref{alg:hetero}). Each
of the above subproblems is solved by a greedy search procedure
(Alg. \ref{alg:greedy}) that starts with the complete conditional MRF
(or MRF), and deletes the edge that yields the greatest improvement to
the PBIC score on each iteration.

\begin{algorithm}[t]
\caption{\textproc{Heterogenous}$(\gee^{\textrm{complete}}, \bface{D})$} \label{alg:hetero}
\begin{algorithmic}[1]
\State $\gee^\bface L, \gee^\bface A, \gee^\bface Y \gets$
conditional MRFs on $\bface L$, $\bface A$, and $\bface Y$ formed
from $\gee ^{\textrm{complete}}$
\State $\cal{E}^*_{\cal{N}_\bface L} \gets \textproc{Greedy Network Search}(\gee^{\bface L}, \bface{D})$
\State $\cal{E}^*_{\cal{N}_\bface A} \gets \textproc{Greedy Network Search}(\gee^{\bface A}, \bface{D})$
\State $\cal{E}^*_{\cal{N}_\bface Y} \gets \textproc{Greedy Network Search}(\gee^{\bface Y}, \bface{D})$
\State \textbf{return} $\cal{E}^*_{\cal{N}_\bface L} \cup \cal{E}^*_{\cal{N}_\bface A}
 \cup \cal{E}^*_{\cal{N}_\bface Y}$
\end{algorithmic}
\end{algorithm}

We now describe procedures for learning network ties in the homogenous
setting, after defining some preliminaries. The \emph{homologs} of an
edge $E_{ij} \in \enetwork$ with endpoints $U_i, W_j \in \bface V$, are
defined as: $h(E_{ij}) \equiv \{ E_{kl} \in \enetwork : \textrm{endpoints}(E_{kl}) =
U_k, W_l\}$. The network tie prototypes in a homogenous graph $\gee$
are defined as: $\eproto \equiv \{E_{ij} \in \mathcal{E}_{ij} \textrm{
	for any } (i, j) \in \mathcal{N} \}$.  $h(\eproto)$ can then be
defined as: $\{h(E) : E \in \eproto\}$.

When the types of connections $\eproto$ between any two connected
units is known, we start with a CG that is fully connected as
$\eproto$ for every pairwise combination of units. Search proceeds by
deleting $\mathcal{E}_{ij}$ between two units $i$ and $j$ that
yields the best improvement in the PBIC on each iteration
(Alg. \ref{alg:homo_known_ties}). When the social network
$\mathcal{N}$ is known, we start with a CG where pairs of units in
$\mathcal{N}$ are fully connected in network ties. Search proceeds by
deleting all homologs of the type of edge in $\eproto$ that yields the
best improvement in the PBIC on each iteration
(Alg. \ref{alg:homo_known_adjs}). Finally, when there is no background
knowledge, homogenous search (Alg. \ref{alg:homo}) can be performed by
chaining the operations of Alg. \ref{alg:homo_known_ties} and Alg. \ref{alg:homo_known_adjs} (or vice versa) on the CG complete
in network ties for every pairwise combination of units.

\begin{algorithm}[t]
\caption{\textproc{Homogenous}$(\gee^{\textrm{complete}}, \bface{D}, \eproto)$} \label{alg:homo_known_ties}
\begin{algorithmic}[1]
\State $\gee^* \gets$ graph obtained by removing all edges between units $i$, $j$ in
$\gee^{\textrm{complete}}$ when $E_{ij} \not\in h(\eproto)$
\State $\textrm{score change} \gets \textrm{True}$
\While{$\textrm{score change}$}
\State $\textrm{score change} \gets \textrm{False}$
\State $\cal{N}^* \gets$ network in $\gee^*$
\State $(i, j)_{max} \gets \argmax_{(i, j) \in \cal{N}^*} \textrm{PBIC}(\bface{D}; \gee^* \setminus \mathcal{E}_{ij})$
\If{ $ \textrm{PBIC}(\bface{D}; \gee^* \setminus \mathcal{E}_{ij_{max}}) > \textrm{PBIC}(\bface{D}; \gee^*)$ }
\State $\gee^* \gets \gee^* \setminus \mathcal{E}_{ij_{max}}$
\State $\textrm{score change} \gets \textrm{True}$
\EndIf
\EndWhile
\State \textbf{return} $\cal{N}^*$
\end{algorithmic}
\end{algorithm}

\begin{algorithm}[t]
\caption{\textproc{Homogenous}$(\gee^{\textrm{complete}}, \bface{D}, \mathcal{N})$} \label{alg:homo_known_adjs}
\begin{algorithmic}[1]
\State $\gee^* \gets $ graph obtained by removing all edges between units $i$, $j$
 in $\gee^{\textrm{complete}}$ when $(i, j) \not\in \cal N$
\State $\textrm{score change} \gets \textrm{True}$
\While{$\textrm{score change}$}
\State $\textrm{score change} \gets \textrm{False}$
\State $\eproto^* \gets$ prototypes of network ties in $\gee^*$
\State $E_{max} \gets \argmax_{E \in \eproto^*} \textrm{PBIC}(\bface{D}; \gee^* \setminus h(E))$
\If{ $\textrm{PBIC}(\bface{D}; \gee^* \setminus h(E_{max}) > \textrm{PBIC}(\bface{D}; \gee^*)$}
\State $\gee^* \gets \gee^* \setminus h(E_{max})$
\State $\textrm{score change} \gets \textrm{True}$
\EndIf
\EndWhile
\State \textbf{return} $\eproto^*$
\end{algorithmic}
\end{algorithm}

\begin{algorithm}[t]
	\caption{\textproc{Homogenous}$(\gee^{\textrm{complete}}, \bface{D})$} \label{alg:homo}
	\begin{algorithmic}[1]
		\State $\eproto \gets$ prototypes of network ties in $\gee^{\textrm{complete}}$
		\State $\cal{N}^* \gets \textproc{Homogenous}(\gee^{\textrm{complete}}, \bface{D}, \eproto)$
		\State $\eproto^* \gets \textproc{Homogenous}(\gee^{\textrm{complete}}, \bface{D}, \cal{N}^*)$
		
		\State \textbf{return} $\cal{N}^*$, $\eproto^*$
	\end{algorithmic}
\end{algorithm}

Clearly we could use the heterogenous procedure even if the true underlying network ties are homogenous, 
since it is most general.
However,
intuitively we expect the homogenous procedures to fare
better in a finite data setting, because the homogeneity assumption
allows pooling data from samples across units for each edge deletion
test. 
This intuition is confirmed in our simulations.

\subsection{Size of the Search Space}

In the heterogenous case, the search
space grows as $O(|\eproto|{m\choose 2})$ i.e., as a function of the
number of possible edges between two units $i$ and $j$ multiplied by
the number of possible pairings on $m$ units. Under homogeneity when
$\eproto$ is known, this reduces to $O({m\choose 2})$; when
$\mathcal{N}$ is known, it reduces to $O(|\eproto|)$; and under
homogeneity where neither is available, it is $O({m\choose 2}) +
O(|\eproto|)$.

\subsection{Consistency of Network Search}

\begin{lemma}
If the generating distribution is Markov to a CG satisfying tier
symmetry and the causal ordering assumption, then the search space of
\textproc{Greedy Network Search} consists of graphs belonging to their
own equivalence classes of size 1.
\label{equivalence_size}
\end{lemma}

\begin{theorem}
\label{gns_consistency}
If the generating distribution is in the exponential family (with compact natural parameter space $\Theta$) and is Markov and faithful to a CG satisfying tier symmetry and causal ordering, then 
\textproc{Greedy Network Search} is consistent.
\end{theorem}


Under the same assumptions in the theorem above, we have the following corollary results.
\begin{corollary}
\label{heterogns_consistency}
The \textproc{Heterogenous} procedure is consistent.
\end{corollary}

\begin{corollary}
\label{homogns_consistency}
When the true network ties are homogenous, the \textproc{Homogenous} procedure is consistent.
\end{corollary}
%

\section{EXPERIMENTS}

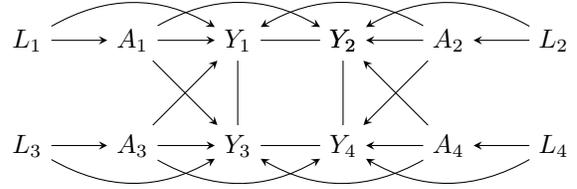
\begin{figure}[H]
	\begin{center}
		\scalebox{1}{
			\begin{tikzpicture}[>=stealth, node distance=1.4cm]
			\tikzstyle{format} = [thick, circle, minimum size=1.0mm, inner sep=0pt]
			\tikzstyle{square} = [draw, thick, minimum size=1.0mm, inner sep=3pt]
			\begin{scope}
			\path[->]
			node[] (l1) {$L_1$}
			node[right of=l1] (a1) {$A_1$}
			node[right of=a1] (y1) {$Y_1$}
			node[right of=y1] (y2) {$Y_2$}
			node[right of=y1] (y2) {$Y_2$}
			node[right of=y2] (a2) {$A_2$}
			node[right of=a2] (l2) {$L_2$}
			node[below of=l1] (l3) {$L_3$}
			node[right of=l3] (a3) {$A_3$}
			node[right of=a3] (y3) {$Y_3$}
			node[right of=y3] (y4) {$Y_4$}
			node[right of=y4] (a4) {$A_4$}
			node[right of=a4] (l4) {$L_4$}
			
			(l1) edge[] (a1)
			(a1) edge[] (y1)
			(l1) edge[bend left] (y1)
			(l2) edge[] (a2)
			(a2) edge[] (y2)
			(l2) edge[bend right] (y2)
			(l3) edge[] (a3)
			(a3) edge[] (y3)
			(l3) edge[bend right] (y3)
			(l4) edge[] (a4)
			(a4) edge[] (y4)
			(l4) edge[bend left] (y4)
			(y1) edge[-] (y2)
			(y2) edge[-] (y4)
			(y4) edge[-] (y3)
			(y3) edge[-] (y1)
			(a1) edge (y3)
			(a3) edge (y1)
			(a2) edge (y4)
			(a4) edge (y2)
			(a1) edge[bend left] (y2)
			(a2) edge[bend right] (y1)
			(a3) edge[bend right] (y4)
			(a4) edge[bend left] (y3)
			
			;
			\end{scope}
			\end{tikzpicture}
		}
	\end{center}
	\caption{The 2-regular CG for a block of size 4}
	\label{fig:cg2}
\end{figure}

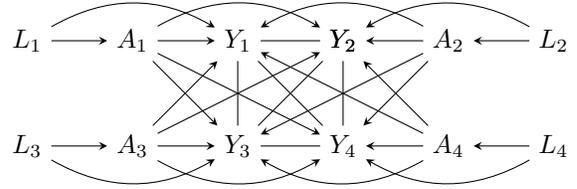
\begin{figure}[H]
	\begin{center}
		\scalebox{1}{
			\begin{tikzpicture}[>=stealth, node distance=1.4cm]
			\tikzstyle{format} = [thick, circle, minimum size=1.0mm, inner sep=0pt]
			\tikzstyle{square} = [draw, thick, minimum size=1.0mm, inner sep=3pt]
			\begin{scope}
			\path[->]
			node[] (l1) {$L_1$}
			node[right of=l1] (a1) {$A_1$}
			node[right of=a1] (y1) {$Y_1$}
			node[right of=y1] (y2) {$Y_2$}
			node[right of=y1] (y2) {$Y_2$}
			node[right of=y2] (a2) {$A_2$}
			node[right of=a2] (l2) {$L_2$}
			node[below of=l1] (l3) {$L_3$}
			node[right of=l3] (a3) {$A_3$}
			node[right of=a3] (y3) {$Y_3$}
			node[right of=y3] (y4) {$Y_4$}
			node[right of=y4] (a4) {$A_4$}
			node[right of=a4] (l4) {$L_4$}
			
			(l1) edge[] (a1)
			(a1) edge[] (y1)
			(l1) edge[bend left] (y1)
			(l2) edge[] (a2)
			(a2) edge[] (y2)
			(l2) edge[bend right] (y2)
			(l3) edge[] (a3)
			(a3) edge[] (y3)
			(l3) edge[bend right] (y3)
			(l4) edge[] (a4)
			(a4) edge[] (y4)
			(l4) edge[bend left] (y4)
			(y1) edge[-] (y2)
			(y2) edge[-] (y4)
			(y4) edge[-] (y3)
			(y3) edge[-] (y1)
			(y1) edge[-] (y4)
			(y2) edge[-] (y3)
			(a1) edge (y3)
			(a3) edge (y1)
			(a3) edge[] (y2)
			(a2) edge (y3)
			(a1) edge[] (y4)
			(a4) edge (y1)
			(a2) edge (y4)
			(a4) edge (y2)
			(a1) edge[bend left] (y2)
			(a2) edge[bend right] (y1)
			(a3) edge[bend right] (y4)
			(a4) edge[bend left] (y3)
			
			;
			\end{scope}
			\end{tikzpicture}
		}
	\end{center}
	\caption{The 3-regular CG for a block of size 4}
	\label{fig:cg3}
\end{figure}

\begin{figure*}[t]
	\centering
	\begin{subfigure}[b]{0.48\textwidth}
		\includegraphics[width=\textwidth]{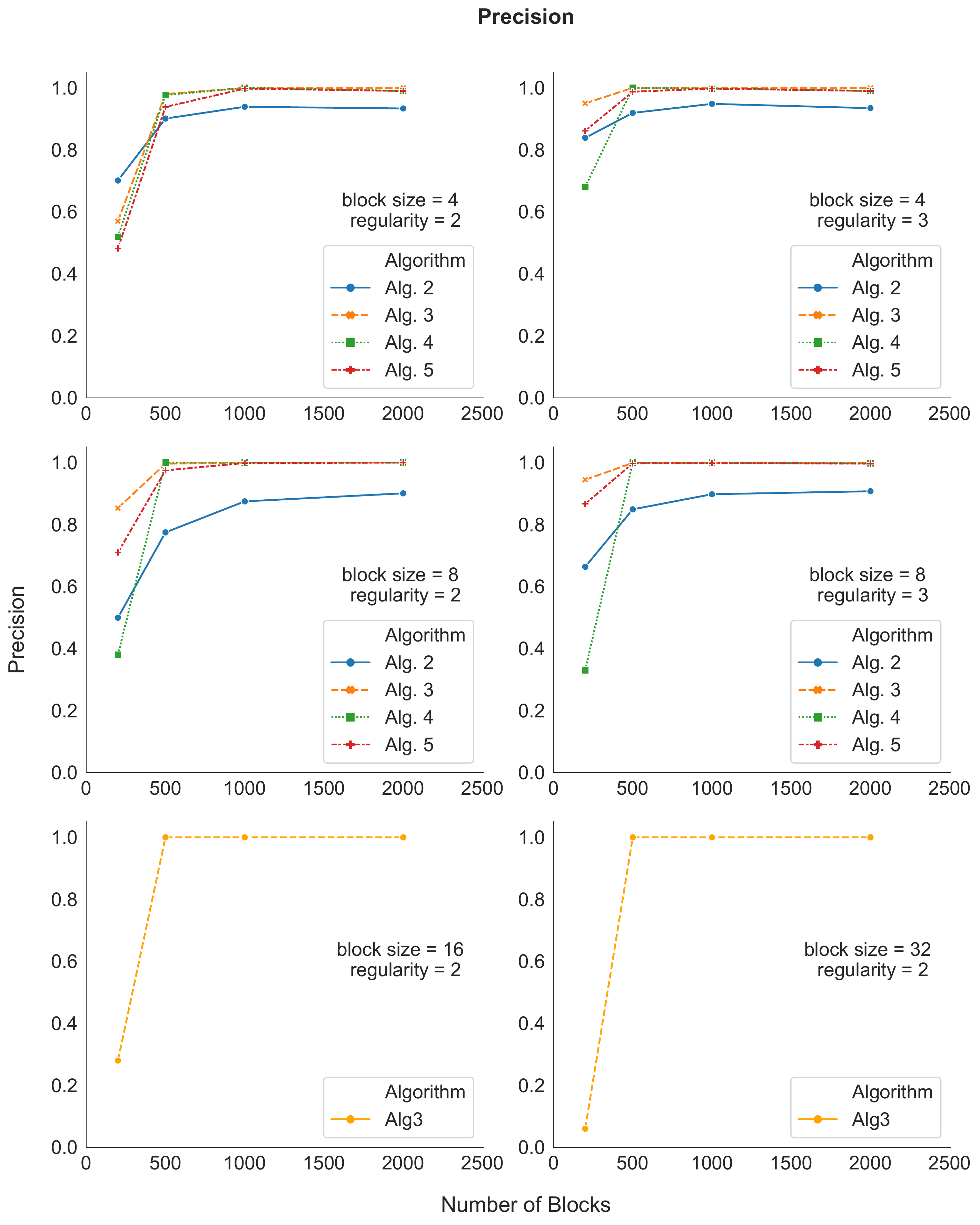}
		\label{precision}
	\end{subfigure}
	\hfill 
	\begin{subfigure}[b]{0.48\textwidth}
		\includegraphics[width=\textwidth]{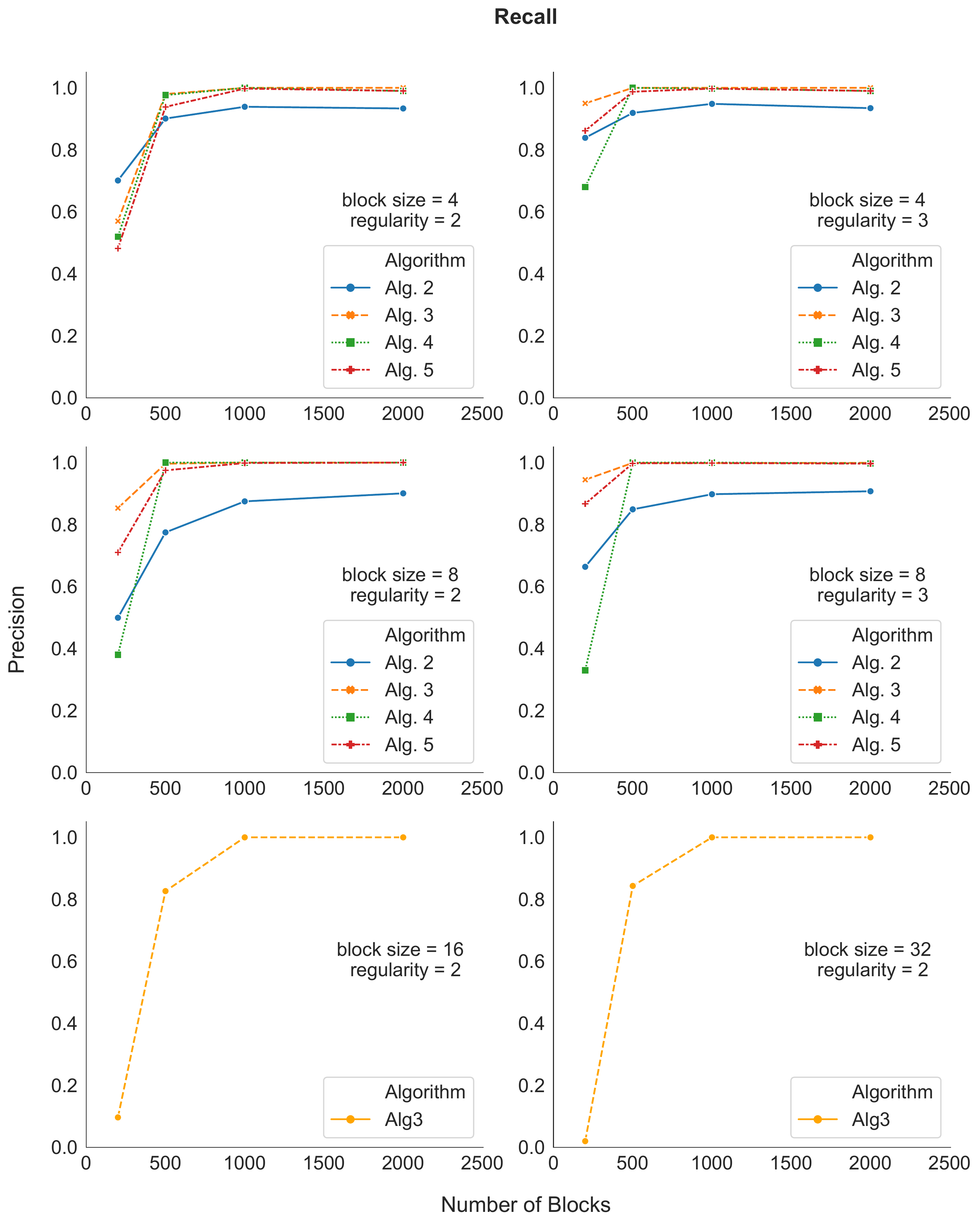}
		\label{recall}
	\end{subfigure}
	\caption{Performance of structure learning algorithms as measured by precision and recall}\label{struct_perf}
\end{figure*}

We evaluate the performance of our proposed algorithms on networks of varying size, for various block sizes, and for different regularity settings. (Regularity refers to the number of neighbors for each unit $i$ in the dependency network ${\cal N}$. This setting thus controls the density of the graph.) We consider blocks of size 4, 8, 16, and 32, with regularity 2 or 3. The ground truth models are homogenous and of the form shown in Figures \ref{fig:cg2} and \ref{fig:cg3}, where we display the case of block size 4.
Data is generated from each network via a Gibbs sampler with a burn-in period of 1000 iterations and thinning every 100 iterations using the following equations:
\begin{align*}
p(L_i=1) &= {\rm expit}(\tau_1),\\
p(A_i=1|L_i,  \{A_j : j \in \nb_{\cal N}(i) \}) &= \\ 
{\rm expit}(\beta_1 L_i + \beta_2 &\sum_{j\in \nb_{\cal N}(i)} A_j), \\ 
p(Y_i=1|L_i, A_i,  \{A_j : j \in \nb_{\cal N}(i) \}) &= \\
{\rm expit}(\nu_1 L_i + \nu_2 A_i + \nu_3 &\sum_{j\in \nb_{\cal N}(i)} A_j),
\end{align*} 
where ${\rm expit}(x)=(1 + {\rm exp}(-x))^{-1}$. We emphasize that some of these networks are quite large; for example, the network with block size 32 and 2000 iid blocks has an effective size of 64,000 individuals. For each network setting we run 100 bootstraps of structure learning in order to get an average estimate of precision and recall as shown in Figure \ref{struct_perf}. However, to spare computation time, we use only Algorithm $\ref{alg:homo_known_ties}$ on the latter two block settings. An interesting feature of the results in Figure \ref{struct_perf}, which matches our earlier intuition, is the faster convergence of the homogenous procedures to the true model -- which we attribute to the parameter sharing (effectively using of more data when testing each edge deletion).

%

In order to demonstrate the utility of learning the structure in
dealing with network uncertainty, we consider the population average overall effect (\ref{eqn:paoe}). We first execute structure learning, and then estimate the PAOE, contrasting a treatment assignment determined with probability 0.7 with the naturally observed probability. 
We do this for 2-regular networks with 2000 realizations of iid blocks of varying size. We use the heterogenous procedure and one of the homogenous procedures (Alg. \ref{alg:homo_known_ties}) to learn the structure of the networks. Estimation
of the causal effect is done by the auto-g-computation algorithm described in \cite{auto-g} and the Supplement. We perform a 1000 bootstraps of both structure learning and effect estimation to compare the bias and variance of the estimates from the learned graphs to the estimates provided by utilizing the maximally uninformative complete graph. Unfortunately the auto-g-computation procedure is also computationally intensive because it requires Gibbs sampling. Again, to spare computation time we do not run the heterogenous procedure on the larger graphs with block sizes 16 and 32 (networks with 32,000 and 64,000 individuals). We also only perform 8 bootstraps for these larger networks. In order to emphasize the need to deal with interference and network uncertainty appropriately, we additionally estimated the bias for 200 bootstraps of the network with blocks of size 8 using the
empty graph (a complete iid assumption), and an incorrect graph where
$\mathcal{N}$ is shuffled randomly to have incorrect adjacencies. In
both cases the bias turned out to be approximately $.06$, an order of magnitude higher than the bias from utilizing the complete or learned graphs.
%

\begin{table}[b]
\begin{tabular}{@{}cccc@{}}
\toprule
\multicolumn{1}{l}{\footnotesize\textbf{Block Size}} & {\footnotesize\textbf{Complete}} & {\footnotesize\textbf{Homogenous}} & {\footnotesize\textbf{Heterogenous}} \\ \midrule
4 & .009, 9.2e-5 & .008, 8.1e-5 & .009, 9.7e-5 \\
8 & .007, 6.6e-5 & .006, 4.1e-5 & .006, 4.5e-5 \\
16 & .006, 3.8e-5 & .005, 1.9e-5 & x \\
32 & .007, 6.1e-5 & .002, 7.6e-6 & x \\ \bottomrule
\end{tabular}
\caption{Bias and variance for estimating the PAOE.}
\label{estimation_tab}
\end{table}

From Table~\ref{estimation_tab} we see that causal effect estimates based on learned structure have the same or lower bias as compared with using the complete graph.  Furthermore, the sparsity of the learned graph reduces
variance of the estimates in most cases. This reduction in bias and
variance is more easily achieved when we are able to exploit
homogeneity in the network structure. In experiments with lower sample sizes, we see that the bias of effect estimates may increase (because the learning procedure may fail to recover the true graph) but that the variance of the estimates remains comparable to or lower than the estimates based on the complete graph.  

\section{CONCLUSION}

We have developed a method for estimating causal effects under unit
dependence induced by a network represented by a chain graph (CG) model
\cite{lauritzen1996}, when there is uncertainty about network structure.
Instead of estimating causal effects given a completely uninformative network
where each pair of units is connected, as is typically done in the interference literature
\cite{tchetgen2012causal, vanderweele2012components},
we estimated causal effects given a sparser network learned via a score-based
model selection method based on the pseudolikelihood function \cite{besag1974}.
We showed that this strategy can yield lower variance in estimates without sacrificing bias, if
the underlying true network structure is recovered accurately.
Our model selection method relied on weak parametric assumptions, specifically that all Markov
factors in the CG model corresponded to conditional Markov random fields in the
exponential family.  The approach here is a generalization of local score-based
search algorithms for directed acyclic graph (DAG) models \cite{chickering2002ges}
to CG models.  As a price of this generalization, our local search algorithms recompute
a potentially larger part of the model score with every move through the model space.
In addition, our approach only works for settings with \emph{partial
interference}, where units within a block exhibit dependence,
but data on blocks is iid. The restriction to blocks of identical size may be relaxed by combining our heterogeneous procedure with a scheme of parameter sharing and hierarchical modeling across blocks that are of different sizes. 
In future work, we aim to extend our methods to \emph{full interference} settings.

\section*{Acknowledgements}
This project is sponsored in part
by the NIH grant R01 AI127271-01 A1, the ONR grant N00014-18-1-2760, and DARPA under contract HR0011-18-C-0049. The content of the information does not necessarily reflect the position or the policy of the Government, and no official endorsement should be inferred. 

\clearpage

{\small
\bibliographystyle{plain}
\bibliography{references}}

\clearpage

\begin{center}
\Large{\bf Supplementary Material}
\end{center}

\section*{CAUSAL CHAIN GRAPHS AND THEIR INTERPRETATION}
Causal models associated with DAGs may be generalized to
causal models associated with CGs.  CGs may include
directed edges, representing direct causation, and undirected edges,
representing symmetric relationships between units in a network.
A causal interpretation of CGs, understood as equilibria of dynamic models with
feedback, was given in \cite{lauritzen2002chain}.
Under this interpretation, the distribution $p(\bface B \mid \pa_\gee(\bface B))$
for each block $\bface B \in \mathcal{B}(\gee)$ can be determined by
a Gibbs sampler on the variables
$B \in \bface B$. Here, each conditional distribution $p(B \mid \bface
B \setminus B, \pa_\gee (\bface B))$ is produced by structural equations
of the form $f_B(\bface B \setminus B, \pa_\gee (\bface B),
\epsilon_B)$. Interventions on elements of $\bface B$ are defined by replacing the appropriate line in the
Gibbs sampler program. For all disjoint sets $\bface Y$ and $\bface A$,
\cite{lauritzen2002chain} showed that $p(\bface Y \mid \doo(\bface a))$
is identified by a CG version of the g-formula \eqref{eqn:cg-g}.

If only interventions on entire blocks are of interest, i.e., we consider
only treatment assignments $\bface A$ such that if $\bface{B} \cap \bface{A} \neq \emptyset$
then $\bface{B} \subseteq \bface{A}$, then an alternative causal interpretation of
a CG ${\cal G}$ that does not rely on the Gibbs sampler machinery of
\cite{lauritzen2002chain} exists.  Specifically, in such a case we
consider a causal DAG model where each block $\bface{B}$ corresponds to
a supervariable $V_\bface{B}$ defined as a Cartesian product of
variables in $\bface{B}$, and a DAG causal model is defined on $V_{\bface{B}}(\bface{A})$, where $\bface{A}$ are values assigned to parents of
$V_\bface{B}$.

If, for each block $\bface{B}$ in a CG ${\cal G}$, the graph $({\cal G}_{\bd_{\cal G}(\bface{B})})^a$ has a single clique, then this yields a classical causal model of a DAG, defined on $\{ V_\bface{B} | \bface{B} \in {\cal B}({\cal G}) \}$.  If not, we can still view the model as a classical causal model of a DAG, but with an extra restriction that the observed data distribution factorizes as (\ref{eqn:cg-f}). See also \cite{ogburn2018causal} for a perspective on interpreting chain graphs in an interference setting.

The model selection methodology introduced here does not depend on which causal interpretation for chain graphs one may choose, and all causal models described above lead to interventional distributions being identified by (\ref{eqn:cg-g}).

\section*{CONDITIONAL MRFs}
A CG model can be viewed as a set of conditional MRFs. A
conditional MRF corresponds to a graph whose vertices can be
partitioned into two disjoint sets: $\bface W$, corresponding to
non-random variables whose values are fixed; and $\bface V$,
corresponding to random variables. The only edges allowed in a
conditional MRF are directed edges $W \rightarrow V$ and undirected
edges $V-V'$\\ for $W \in \bface{W}$ and $V,V' \in \bface{V}$. A statistical model associated with a
conditional MRF $\gee$ is a set of densities that factorize as:
\begin{equation*}
p(\bface{V} \mid \bface{W}) = 
\frac{
	\prod_{ \{\bface{C} \in {\cal C}(({\cal G}_{\bd_{\cal G}(\bface{V})})^a): \bface{C} \not\subseteq \bface{W} \} }
	\phi_\bface{C}(\bface{C})
}{Z({\bf W})}
\notag
\end{equation*}

It is easy to see that the above factorization is analogous to the
second level of CG factorization found in \eqref{eqn:cg-f} where
$\bface{V}$ is a block, and $\bface{W}$ are its parents.

\section*{THE AUTO-G-COMPUTATION ALGORITHM}

The auto-g-computation algorithm, introduced in \cite{auto-g}, may be viewed as a generalization of the Monte Carlo sampling version of the g-computation algorithm for classical causal models (represented by DAGs) \cite{westreich2012parametric} to causal models of the sort we consider here, represented by CGs. We describe a version of this algorithm
based on the pseudolikelihood estimator.
An alternative based on the coding estimator \cite{besag1974} is less efficient, but leads to asymptotically normal estimators of the population average overall effect (PAOE).

Auto-g-computation generates samples from either the observed data distribution that factorizes as (\ref{eqn:cg-f}) according to a CG, or of functions of these distributions, such as counterfactual expectations identified using (\ref{eqn:paoe-id}).

This is done by imposing a topological ordering on blocks in a CG, and generating samples for each block sequentially using Gibbs sampling.  The parameters for Gibbs factors used in the sampler (which by the global Markov property for CGs take the form of
$p(  X_{i} | X_{\bd_{\cal G}(X_i)} )$) are learned via maximizing the pseudolikelihood function.
For any block ${\bf X}$, the Gibbs sampler draws samples from $p(\mathbf{X} \mid \bd_{\cal G}({\bf X}))$, given a fixed set of samples drawn from all blocks with elements in $\pa_{\cal G}({\bf X})$ as follows:

\underline{Gibbs Sampler for ${\bf X}$}:
\begin{align*}
\text{for }t  &  =0,\text{let } \mathbf{x}^{(0)}  \text{ denote initial values ;}\\
\text{for }t  &  =1,...,T\\
&  \text{draw value of }X_{1}^{(t)}\text{ from }p(  X_{1} | {\bf x}^{(t-1)}_{\bd_{\cal G}(X_1)} )
);\\
&  \text{draw value of }X_{2}^{(t)}\text{ from }p(  X_{2} | {\bf x}^{(t-1)}_{\bd_{\cal G}(X_2)} )
);\\
&  \vdots\\
&  \text{draw value of }X_{m}^{(t)}\text{ from }p(  X_{m} | {\bf x}^{(t-1)}_{\bd_{\cal G}(X_m)} )
);
\end{align*}

This method may be used to estimate the counterfactual expectation in (\ref{eqn:paoe-id}) as follows.  We first generate a set of samples
$\bface{L}^{(t)}$, $t = 1, \ldots, T$.  Then we generate a sample $\bface{A}$ directly using some $\pi_i(\bface{A})$, $i = 1, 2$.
Finally, we use the above samples to generate a set of samples $\bface{Y}^{(t)}$, $t = 1, \ldots, T$ using Gibbs factors
$p(Y_i \mid \bface{A}_{\bface{A} \cap \bd_{\cal G}(Y_i)}, \bd_{\cal G}(Y_i) \setminus \bface{A})$.  Finally, we estimate
\[
\frac{1}{m} \sum_{i=1}^m \mathbb{E}[Y_i(\bface{A})] = \frac{1}{m \cdot T}  \sum_{i=1}^m \sum_{t=1}^T Y^{(t)}_i.
\]
It is not difficult to show, (see \cite{auto-g} for details), that rerunning this procedure with different draws $\bface{A}$ from either $\pi_1(\bface{A})$
or $\pi_2(\bface{A})$, and taking the difference of the resulting averages yields a valid estimate of the PAOE.

Fitting parameters of Gibbs factors using the pseudolikelihood function avoids the usual difficulties CGs inherit from Markov random fields, specifically, the intractability of the likelihood function due to the presence of normalizing functions.  In addition, if the learned block structure is sparse, while the number of independent samples considered is small, this approach allows one to impose parameter sharing among Gibbs factors, which leads to reasonable estimates even in small samples.  Taken to the extreme, this approach allows inferences to be made even from a \emph{single sample} of a network, as discussed in detail in \cite{auto-g}.  In this manuscript we only consider the setting where multiple independent samples from blocks are available.

\section*{COMPUTATIONAL COMPLEXITY OF COMPUTING SCORES OF A CHAIN GRAPH MODEL}

In blocks of a CG, the number of local terms that need to be computed
corresponds to the number of vertices present in cliques containing
the edge of interest in the augmented subgraph of the block and its
parents. A term for $V_j$ requires an $O(|\bd_{\cal
	G}(V_j)|)$ computation to update, which in the worst case may be
exponential in the number of vertices if the graph is not sparse.  In
search problems, restrictions can be made on the maximum size of the
boundary set, sacrificing accuracy for tractability.  For a block in a
CG corresponding to a conditional MRF in the exponential family, and
an edge that is present in a set of cliques spanning all vertices, we
will have a local set of size $O(d)$ in the worst case, with each
local term requiring an $O(\textrm{clique size})$ computation.  Thus,
limiting the maximum clique size may speed up the computation of each
local term, but in many cases we may be unable to avoid an $O(d)$
number of such terms.  In other words, our scoring method for CG
models where blocks correspond to conditional MRFs in the exponential
family may not scale to very large graphs, even if such graphs are
sparse.  Achieving such a scaling will entail making additional
assumptions, such as Gaussianity, or non-existence of higher order
interaction terms in log-linear models.  We contrast this with DAG
models, where the local set is of constant size regardless of
parametric assumptions made.

\section*{FORWARD-BACKWARD SEARCH}

Consistency of the score was sufficient to show consistency of a
backwards greedy search involving only edge deletions starting from a
complete conditional MRF. \cite{chickering2002ges} showed that a
property called \emph{local consistency}, which follows from
decomposability and consistency of the score, is sufficient to design a consistent
forward-backward greedy search in the space of (Markov equivalent) DAGs. The forward stepwise
search considers additions, rather than deletions, of single edges to
improve the score, which typically produces a more sparse starting model
for the subsequent backwards search.

Consider a graph $\gee$ and another $\gee'$ that differs only by
the addition of
an edge $V_i - V_j$ or $V_i \rightarrow V_j$. A score $S(\bface D; \gee)$ is called locally consistent if:
\begin{enumerate}
	\item $V_i \not\ci_{\gee_0} V_j \mid \bd_{\gee} (V_i)$ \textbf{or} $V_j
	\not\ci_{\gee_0} V_i \mid \bd_{\gee} (V_j)$ then $\lim_{n\to\infty}P(S(\bface D; \gee') > S(\bface D; \gee)) \to 1$
	\item $V_i \ci_{\gee_0} V_j \mid \bd_{\gee'} (V_i)$ \textbf{and}
	$V_j \ci_{\gee_0} V_i \mid \bd_{\gee}(V_j)$ then $\lim_{n\to\infty}P(S(\bface D; \gee') < S(\bface D; \gee)) \to 1$
	
\end{enumerate}

Such a property requires a stronger notion of decomposability than is
available in our general setting. In Section $4.2$ we mention that if
our model is an MRF that is multivariate normal, or corresponds to a
log linear discrete model with only main effects and pairwise
interactions, then it suffices to consider the following terms derived from the
local set: $\{ s(V_i, \bd_\gee(V_i)), s(V_j, \bd_\gee(V_j)) \}$ for an
edge $V_i - V_j$, and $\{ s(V_j, \bd_\gee(V_j)) \}$ for an edge $V_i
\rightarrow V_j$ (dropping implicit $\bface D$ and $\gee$ for
brevity). This is the strong notion of decomposability we need for
local consistency. Thus, in such settings one can follow the work in
\cite{chickering2002ges} to show that PBIC will be locally consistent
and design a search procedure involving a forward phase followed by a
backward phase. The advantage of such a procedure is that it is more
scalable, even more so when the underlying true model is sparse.

\section*{PROOFS} \label{proofs}

Let $\emm_0$ denote the true model and $\emm_1$, $\emm_2$ two candidate models.
A scoring criterion $S(\bface D; \emm)$ is said to be
\textit{consistent} if:
\begin{align*}
\lim_{n\to\infty} P_n(S(\bface D; \emm_1) &< S(\bface D; \emm_2)) \to 1 \textrm{ when } \\
\emm_1 \not\supseteq \emm_0 &\textrm{ and } \emm_2 \supseteq \emm_0 \textrm{ or} \tag{*} \label{*} \\
\emm_1, \emm_2 \supseteq \emm_0 &\textrm{ and } k_1 > k_2. \tag{**} \label{**} 
\end{align*}

\begin{lema}{\ref{pbic_consistency}}
	With dimension fixed and sample size increasing to infinity,
	the PBIC is a consistent score for curved exponential families whose natural parameter space $\Theta$ forms a compact set.
\end{lema}

\begin{proof}

	To prove consistency we need to show that,
	\begin{equation}
	\lim_{n\to\infty} P_n(PBIC(\bface D; \emm_1) < PBIC(\bface D; \emm_2)) \to 1
	\label{consistency}
	\end{equation}
	when \eqref{*} or \eqref{**}.
	
	Note in all following steps, we assume $\bface D$ to be implicit in the calculation
	of the likelihoods and pseudolikelihoods.
	
	To prove \eqref{consistency} holds under the scenario \eqref{*}, it is sufficient
	to show that the following is true for some $\epsilon > 0$
	\begin{equation}
	\frac{1}{n} (\ln \mathcal{PL}_n(\hat{\theta}_2) - \ln \mathcal{PL}_n(\hat{\theta}_1)) > \epsilon
	\label{positive_pl_diff}
	\end{equation}
	
	It was shown in \cite{haughton1988} that for any $\emm_1$ outside of a
	neighbourhood $N$ of $\theta_0$, and $\emm_2$ containing this
	neighbourhood, we can pick a $\delta > 0$ such that:
	\begin{equation}
	\frac{1}{n} (\ln \mathcal{L}_n(\hat{\theta}_2) - \ln \mathcal{L}_n(\hat{\theta}_1)) > \delta
	\label{positive_l_diff}
	\end{equation}
	
	In order to extend this result to \eqref{positive_pl_diff}, we
	invoke a result from \cite{mozeika2014} stating that
	\begin{equation}
	\mathcal{PL}_n(\theta) \geq d \mathcal{L}_n(\theta) + \sum_{i=1}^{d}H_i(\widetilde{P}_n)
	\label{pl_l_ineq}
	\end{equation}
	where $d$ is the dimensionality of the data, and $H_i(\widetilde{P}_n)$ is the Shannon entropy of the empirical
	distribution. It then follows that \eqref{positive_pl_diff} holds when
	\eqref{positive_l_diff} is true.
	
	Showing that \eqref{consistency} holds under the scenario \eqref{**}
	is equivalent to showing that the following difference is $O_p(1/n)$:
	\begin{equation}
	\frac{1}{n}|\ln \mathcal{PL}_n(\hat{\theta}_{1}) - \ln \mathcal{PL}_n(\hat{\theta}_{2})|
	\label{pl_diff}
	\end{equation}
	
	Consider the difference between the full log-likelihoods:
	\begin{equation}
	\frac{1}{n}|\ln \mathcal{L}_n(\hat{\theta}_{1}) - \ln \mathcal{L}_n(\hat{\theta}_{2})|.
	\label{l_diff}
	\end{equation}
	
	We first closely follow the proof in \cite{haughton1988} to show that the quantity in \eqref{l_diff} is $O_p(1/n)$. Consider data drawn from a curved exponential family density $p({\bf X}; \theta)=h({\bf X}){\rm exp}(\theta T({\bf X}) - Z(\theta))$, where $\theta \in \mathbb{R}^k$ is a set of canonical parameters in the natural parameter space $\Theta$, $T({\bf X})$ is a set of sufficient statistics, and $Z(\theta)$ is a normalizing function. For a particular choice of a model ${\cal M}$ in this setting, the BIC can be written as ${\rm ln}\mathcal{L}_n({\bf D}; \hat{\theta}) - \frac{k}{2}{\rm ln}(n)$ or equivalently,
	\begin{equation}
	\sup_{\theta \in {\cal M} \cap \Theta} \sum_{i=1}^{n}{\rm }\theta T({\bf X}_i) - Z(\theta) - \frac{k}{2}{\rm ln}(n),
	\label{haughton_bic}
	\end{equation}

	Note that for simplicity of notation and without loss of generality, we set $h({\bf X})=1$. Now consider ${\bf T}_n = \frac{1}{n}\sum_{i=1}^{n}T({\bf X}_i)$, the sample average of the sufficient statistics. We can then express \eqref{haughton_bic} as
	\begin{equation}
	n \sup_{\theta \in {\cal M} \cap \Theta} \theta {\bf T}_n - Z(\theta) - \frac{k}{2}{\rm ln}(n).
	\label{sample_bic}
	\end{equation}
	
	Define the quantities $S_{n, i}$ and $U_n$ as,
	\begin{align*}
	S_{n, i} &\equiv \sup_{\theta_i \in {\cal M}_i \cap \Theta} \theta_i {\bf T}_n - Z(\theta_i) = \hat{\theta}_{n, i} {\bf T}_n - Z(\hat{\theta}_{n, i}),\\
	U_n &\equiv \theta_0{\bf T}_n - Z(\theta_0),
	\end{align*}
	
	where $\hat{\theta}_{n, i}$ is the MLE. 
	We now show that $S_{n, i} - U_n$ and by extension each term in \eqref{l_diff} is $O_p(1/n)$. Since $\theta_0$ lies in both model spaces under scenario \eqref{**},
	\begin{equation}
	S_{n, i} - U_n = (\hat{\theta}_{n, i} - \theta_0) {\bf T}_n - Z(\hat{\theta}_{n, i}) + Z(\theta_0) \geq 0.
	\label{score_diff}
	\end{equation}
	
	Considering the Taylor expansion of $Z$ about $\theta_0$, we have that $Z(\hat{\theta}_{n, i}) - Z(\theta_0) = (\hat{\theta}_{n, i} - \theta_0)\nabla Z(\theta_0) + O_p(1/n)$, where the $O_p(1/n)$ term comes from the efficiency of MLE \cite{huber1967}. Plugging this into \eqref{score_diff} we get,
	\begin{equation}
	S_{n, i} - U_n = ({\bf T}_n - \nabla Z(\theta_0))(\hat{\theta}_{n, i} - \theta_0)  + O_p(1/n).
	\end{equation}
	By the Central Limit Theorem, ${\bf T}_n - \nabla Z(\theta_0)$ is $O_p(1/\sqrt{n})$ and by the efficiency of MLE, $\hat{\theta}_{n, i} - \theta_0$ is also $O_p(1/\sqrt{n})$. Thus, $S_{n, i} - U_n$ is $O_p(1/n)$, and we have our result.

	In order to extend this result to \eqref{pl_diff}, we once again
	invoke the result from \cite{mozeika2014} that
	\begin{equation}
	\mathcal{PL}_n(\theta) \geq d \mathcal{L}_n(\theta) + \sum_{i=1}^{d}H_i(\widetilde{P}_n)
	\end{equation}
	where $H_i(\widetilde{P}_n)$ is the Shannon entropy of the empirical
	distribution. We see that as long $d \ll n$ (which in our
	setting we assume to be true), \eqref{l_diff} being $O_p(1/n)$ implies
	that \eqref{pl_diff} is as well.
\end{proof}

\begin{lema}{\ref{locality}}
	Let $\gee$ and $\gee'$ be graphs which differ by a single edge between $V_i$ and $V_j$.
	For conditional MRFs in the exponential family, the local score
	difference between $\gee$ and $\gee'$ is given by:
	$\sum_{V \in \loc(V_i, V_j; \gee) \cap \bface{B}_{\loc}} \{s_V\big(\bface{D}; \gee \big) - s_V\big(\bface{D}; \gee' \big)\},$ where
	$s_V(.)$ denotes the component of the score for $V$.
\end{lema}

\begin{proof}
	A conditional MRF corresponding to $p(\bface{B} \mid \pa_{\cal G}(\bface{B}))$ for a block $\bface{B}$ in a CG
	${\cal G}$ in the (conditional) exponential family has a probability
	distribution of the general form:
	\begin{align}
	&p(\bface{B} \mid \pa_{\cal G}(\bface{B}); \psi) =\\
	\notag
	&\textrm{exp}\left(
	\sum_{ \{ \bface{C} \in {\cal C}( ({\cal G}_{\bd_{\cal G}(\bface{B})})^a ): \bface{C} \not\subseteq \pa_{\cal G}(\bface{B}) \} }
	\!\!\!\!\! \psi_\bface{C} T(\bface{C}) - Z(\psi, \pa_{\cal G}(\bface{B}))
	\right)
	\end{align}
	where
	\[
	\left\{ \psi_\bface{C} : \bface{C} \in {\cal C}( ({\cal G}_{\bd_{\cal G}(\bface{B})})^a ), \bface{C} \not\subseteq \pa_{\cal G}(\bface{B}) \right\}
	\]
	is a set of canonical parameters associated with potential functions $\phi_\bface{C}$ in the CG factorization,
	\[
	\left\{ T(\bface{C}) : \bface{C} \in {\cal C}( ({\cal G}_{\bd_{\cal G}(\bface{B})})^a ), \bface{C} \not\subseteq \pa_{\cal G}(\bface{B}) \right\}
	\]
	is a set of sufficient statistics for $\psi_\bface{C}$,
	and $Z(\theta,\pa_{\cal G}(\bface{B}))$ is a normalizing function.
	
	Assume $V$ is in a clique $\bface{C}$ that contains the edge $V_i - V_j$
	in ${\cal G}$, and let ${\cal G}^{-}$ be the edge subgraph of ${\cal
		G}$ with that edge removed.  Then $p(V \mid \bd_{\cal G}(V))$ will
	only be a function of clique parameters $\psi_{\bf S}$, where ${\bf S}
	\subseteq {\cal C}( ({\cal G}_{\bd_{\cal G}(\bface{B})})^a ): \bface{C}
	\not\subseteq \pa_{\cal G}(\bface{B})$ and $V \in \bface S$. All others
	terms in the factorization cancel by definition of conditioning.
	As a consequence, 
	$p(V \mid \bd_{\cal G}(V))$  will be a function of $\psi_\bface{C}$.
	
	However, after $V_i - V_j$ is removed, $\bface{C}$ will no longer be a
	clique in ${\cal G}^{-}$, by definition, but will instead decompose
	into two cliques, say $\bface{C}_1$ and $\bface{C}_2$.  By following the
	above reasoning, $p(V \mid \bd_{{\cal G}^{-}}(V))$ will be a function
	of all clique parameters $\{ \psi_{\bf S}$ : ${\bf S}
	\subseteq {\cal C}( ({\cal G}_{\bd_{\cal G}(\bface{B})})^a ), \bface{C}
	\not\subseteq \pa_{\cal G}(\bface{B}), V \in \bface S \}$,  which will include
	$\psi_{\bface{C}_1}$ and $\psi_{\bface{C}_2}$.  Since the parameterization
	for $p(V \mid \bd_{{\cal G}^-}(V))$ is thus different in models for
	${\cal G}$ and ${\cal G}^{-}$, the contribution to the score
	associated with this term will also be different.
	
	Assume $V$ is not in a clique that contains the edge $V_i - V_j$ in ${\cal G}$, and let
	${\cal G}^{-}$ be the edge subgraph of ${\cal G}$ with that edge removed, as before.
	Then $p(V \mid \bd_{\cal G}(V))$ will only be a function of clique parameters $\psi_{\bf S}$,
	where ${\bf S}$ contains $V$, all others will cancel by definition of conditioning.
	
	Note that since no such ${\bf S}$ contains the edge $V_i - V_j$ in
	${\cal G}$, the set of cliques $\bface S$ in $\gee$ is the same as the
	set of cliques $\bface S$ in $\gee^-$. Moreover, since ${\cal G}^-$ is
	an edge subgraph of ${\cal G}$, no new cliques are introduced.  As a
	result, $p(V \mid \bd_{{\cal G}^-}(V))$ will be parameterized by the
	same set of $\psi_{\bf S}$ in the model for ${\cal G}^-$ as it was in
	the model for ${\cal G}$.
	
	Our conclusion then follows because, by properties of the exponential
	family, the sufficient statistics for a clique parameter $\psi_{\bf
		S}$ are functions of only $\bface S$.  Since draws from $p({\bf S})$
	are fixed, the estimates for $\psi_{\bf S}$ will coincide if the data
	is evaluated under the model for ${\cal G}$, and the model for ${\cal
		G}^-$.  Furthermore, the number of parameters in $p(V \mid
	\bd_{{\cal G}}(V))$ and $p(V \mid \bd_{{\cal G}^-}(V))$ is the same.
	This implies the score contribution for $p(V \mid \bd_{\cal G}(V))$ in
	${\cal G}$ will equal the score contribution of $p(V \mid \bd_{{\cal
			G}^-}(V))$ in ${\cal G}^-$. The only terms remaining in the score
	difference between $\gee$ and $\gee'$ are then local scores for $V \in
	\loc(V_i, V_j; \gee)$.
	
	This implies the conclusion.
\end{proof}

\begin{lema}{\ref{equivalence_size}}
	If the generating distribution is Markov to a CG satisfying tier
	symmetry and the causal ordering assumption, then the search space of
	\textproc{Greedy Network Search} consists of graphs belonging to their
	own equivalence classes of size 1.
\end{lema}

\begin{proof}
	Under the restrictions listed above, the only changes allowed are edge
	deletions or additions of the form $L_i - L_j$, $A_i - A_j$, $Y_i -
	Y_j$, $L_i \rightarrow A_j$, $L_i \rightarrow Y_j$, $A_i \rightarrow
	Y_j$.
	
	Consider an edge deletion $V_i - V_j$ in $\gee$, giving rise to a
	graph $\gee'$. Notice that boundaries of $V_i$ and $V_j$ have
	changed. Thus by the local Markov property on chain graphs, $\gee$ and
	$\gee'$ must imply different conditional independences. Concretely,
	$\gee$ implies:
	\begin{align*}
	V_i &\ci \bface V \setminus \cl_{\gee}(V_i) \mid \bd_{\gee}(V_i) \\
	V_j &\ci \bface V \setminus \cl_{\gee}(V_j) \mid \bd_{\gee}(V_j) \\
	\end{align*}
	
	while $\gee'$ implies:
	\begin{align*}
	V_i &\ci \bface V \setminus (\cl_{\gee}(V_i) \setminus V_j) \mid \bd_{\gee}(V_i) \setminus V_j \\
	V_j &\ci \bface V \setminus (\cl_{\gee}(V_j) \setminus V_i) \mid \bd_{\gee}(V_j) \setminus V_i \\
	\end{align*}
	
	We can similarly show that an edge deletion $V_i \rightarrow V_j$ also
	implies different conditional independences in $\gee$ and
	$\gee'$. Thus, in general, an edge deletion or addition in our search
	space gives rise to graphs that are not Markov equivalent and hence,
	reside in their own equivalence classes of size 1.
	
\end{proof}

\begin{thma}{\ref{gns_consistency}}
	If the generating distribution is in the exponential family (with compact natural parameter space $\Theta$) and is Markov and faithful to a CG satisfying tier symmetry and causal ordering, then 
	\textproc{Greedy Network Search} is consistent.
\end{thma}

\begin{proof}
	The algorithm begins with a complete conditional MRF that contains the
	true underlying distribution.  We are guaranteed that the truth is
	contained in every state through the entirety of the algorithm by the
	following argument.  Consider the first edge deletion performed by GNS
	to a conditional MRF that does not contain the true model. It follows
	from consistency of the PBIC that any such deletion would decrease the
	score. Choosing such an edge deletion would contradict the greediness
	of the algorithm.
	
	Now assume the algorithm stops at a sub optimal conditional MRF $\gee$
	that contains the truth but has more parameters than the true model
	$\gee^*$. We know there exists a series of single edge deletions in
	$\enetwork$ that takes us from $\gee$ to $\gee^*$. By Lemma
	\ref{equivalence_size}, each of these edge deletions yield graphs in
	separate equivalence classes. It follows then from the consistency of
	the PBIC that each of these edge deletions strictly increases the
	score (each edge deletion yields a smaller model containing the truth)
	and thus, a local optimum found by greedily maximizing the PBIC
	corresponds to finding the global optimum $\gee^*$.
\end{proof}

\begin{cora}{\ref{heterogns_consistency}}
	The \textproc{Heterogenous} procedure is consistent.
\end{cora}

\begin{proof}
	By consistency of GNS, each conditional MRF returned for $\bface L$, $\bface A$, and
	$\bface Y$ corresponds to the true model. The union of these will then produce the true
	CG on $\bface V$.
\end{proof}

\begin{cora}{\ref{homogns_consistency}}
	When the true network ties are homogenous, \textproc{Homogenous} network search is consistent.
\end{cora}

\begin{proof}
	Each of the homogenous procedures described above can be decomposed
	into a series of single edge deletions that we have shown to be
	consistent.
\end{proof}

\end{document}